%% file: main.tex
\documentclass{article}

\usepackage{microtype}
\usepackage{graphicx}
\usepackage{subfigure}
\usepackage{booktabs} 
\usepackage{amssymb}
\usepackage{amsmath}
\usepackage{float}
\usepackage{multirow}
\usepackage{hyperref}

\usepackage[accepted]{icml2019}

\icmltitlerunning{Self-Attention Graph Pooling}

\begin{document}

\twocolumn[
\icmltitle{Self-Attention Graph Pooling}



\icmlsetsymbol{equal}{*}

\begin{icmlauthorlist}
\icmlauthor{Junhyun Lee}{equal,ku}
\icmlauthor{Inyeop Lee}{equal,ku}
\icmlauthor{Jaewoo Kang}{ku}
\end{icmlauthorlist}

\icmlaffiliation{ku}{Department of Computer Science and Engineering, Korea University, Seoul, Korea}

\icmlcorrespondingauthor{Jaewoo Kang}{kangj@korea.ac.kr}
\icmlkeywords{Deep learning, Graph pooling, Graph neural network, Architecture}
\vskip 0.3in
]



\printAffiliationsAndNotice{\icmlEqualContribution{}} 

\begin{abstract}
Advanced methods of applying deep learning to structured data such as graphs have been proposed in recent years. In particular, studies have focused on generalizing convolutional neural networks to graph data, which includes redefining the convolution and the downsampling (pooling) operations for graphs. The method of generalizing the convolution operation to graphs has been proven to improve performance and is widely used. However, the method of applying downsampling to graphs is still difficult to perform and has room for improvement. In this paper, we propose a graph pooling method based on self-attention. Self-attention using graph convolution allows our pooling method to consider both node features and graph topology. To ensure a fair comparison, the same training procedures and model architectures were used for the existing pooling methods and our method. The experimental results demonstrate that our method achieves superior graph classification performance on the benchmark datasets using a reasonable number of parameters.
\end{abstract}

\input{sections/1_introduction.tex}
\input{sections/2_related_work.tex}

\input{sections/3_proposed_method.tex}
\input{sections/4_experiments.tex}

\input{sections/5_analysis.tex}

\input{sections/6_conclusion.tex}

\bibliographystyle{icml2019}
\end{document}

%% file: sections/1_introduction.tex
\section{Introduction}
\label{introduction}
The advent of deep learning has led to extensive improvements in technology used to recognize and utilize patterns in data \cite{lecun2015deep}. In particular, convolutional neural networks (CNNs) successfully leverage the properties of data such as images, speech, and video on Euclidean domains (grid structure)  \cite{hinton2012deep, krizhevsky2012imagenet, he2016deep, karpathy2014large}. CNNs consist of convolutional layers and downsampling (pooling) layers. The convolutional and pooling layers exploit the shift-invariance (also known as stationary) property and compositionality of grid-structured data \cite{simoncelli2001natural, bronstein2017geometric}. As a result, CNNs perform well with a small number of parameters.

In various fields, however, a large amount of data, such as graphs, exists on the non-Euclidean domain. For example, social networks, biological networks, and molecular structures can be represented by nodes and edges of graphs \cite{lazer2009life,davidson2002genomic,duvenaud2015convolutional}. Therefore, attempts have been made to successfully use CNNs in the non-Euclidean domain. Most previous studies have redefined the convolution and pooling layers to process graph data.

To define graph convolution, studies have used the spectral \cite{bruna2014spectral,henaff2015deep,defferrard2016convolutional,kipf2016semi} and non-spectral  \cite{monti2017geometric,hamilton2017inductive,Xu2018RepresentationLO,veličković2018graph,DBLP:journals/corr/abs-1810-02244} methods. The application of graph convolution has resulted in outstanding performance in a variety of fields which include recommender systems \cite{van2017graph, yao2018convolutional,monti2017geometric}, chemical researches \cite{NIPS2018_7877, Zitnik2018}, natural language processing \cite{bastings2017graph, peng2018large, yao2018graph}, and in many tasks as reported in \citeauthor{zhou2018graph}.

There are fewer methods for graph pooling than for graph convolution. Previous researches have adopted the pooling method that considers only graph topology \cite{defferrard2016convolutional, ijcai2018-490}. With growing interest in graph pooling, several improved methods have been proposed \cite{dai2016discriminative, duvenaud2015convolutional,gilmer2017neural,zhang2018end}. They utilize node features to obtain a smaller graph representation. Recently, \citeauthor{RexYing,gao2019graph,cangea2018towards} have proposed innovative pooling methods that can learn hierarchical representations of graphs. These methods allow Graph Neural Networks (GNNs) to attain scaled-down graphs after pooling in an end-to-end fashion.

However, the above pooling methods have room for improvement. For example, the differentiable hierarchical pooling method of \citeauthor{RexYing} has a quadratic storage complexity and the number of its parameters is dependent on the number of nodes. \citeauthor{gao2019graph,cangea2018towards} have addressed the complexity issue, but their method does not take graph topology into account.

Here, we propose SAGPool which is a Self-Attention Graph Pooling method for GNNs in the context of hierarchical graph pooling. Our method can learn hierarchical representations in an end-to-end fashion using relatively few parameters. The self-attention mechanism is exploited to distinguish between the nodes that should be dropped and the nodes that should be retained. 
Due to the self-attention mechanism which uses graph convolution to calculate attention scores, node features and graph topology are considered.
In short, SAGPool, which has the advantages of the previous methods, is the first method to use self-attention for graph pooling and achieve high performance.
The code is available on Github \footnote{\url{https://github.com/inyeoplee77/SAGPool}}

%% file: sections/2_related_work.tex
\section{Related Work}
\label{related}
GNNs have drawn considerable attention due to their state-of-the-art performance on tasks in the graph domain. Studies on GNNs focus on extending the convolution and pooling operation, which are the main components of CNN, to graphs. 

\subsection{Graph Convolution} 
\label{related:graphconvolution}
Convolution operation on graphs can be defined in either the spectral or non-spectral domain. Spectral approaches focus on redefining the convolution operation in the Fourier domain, utilizing spectral filters that use the graph Laplacian. \citeauthor{kipf2016semi} proposed a layer-wise propagation rule that simplifies the approximation of the graph Laplacian using the Chebyshev expansion method \cite{defferrard2016convolutional}.
The goal of non-spectral approaches is to define the convolution operation so that it works directly on graphs. In general non-spectral approaches, the central node aggregates features from adjacent nodes when its features are passed to the next layer rather than defining the convolution operation in the Fourier domain. \citeauthor{hamilton2017inductive} proposed GraphSAGE which learns node embeddings through sampling and aggregation. While GraphSAGE operates in a fixed-size neighborhood, Graph Attention Network (GAT) \cite{veličković2018graph}, based on attention mechanisms \cite{DBLP:journals/corr/BahdanauCB14}, computes node representations in entire neighborhoods. Both approaches have improved performance on graph-related tasks. 

\subsection{Graph Pooling}
Pooling layers enable CNN models to reduce the number of parameters by scaling down the size of representations, and thus avoid overfitting. To generalize CNNs, the pooling method for GNNs is necessary. Graph pooling methods can be grouped into the following three categories: topology based, global, and hierarchical pooling.

\textbf{Topology based pooling}
Earlier works used graph coarsening algorithms rather than neural networks. Spectral clustering algorithms use eigendecomposition to obtain coarsened graphs. However, alternatives were needed due to the time complexity of eigendecomposition. Graclus\cite{graclus} computes clustered versions of given graphs without eigenvectors because of the mathematical equivalence between a general spectral clustering objective and a weighted kernel k-means objective. Even in recent GNN models \cite{defferrard2016convolutional,ijcai2018-490}, Graclus is employed as a pooling module.


\textbf{Global pooling}
Unlike the previous methods, global pooling methods consider graph features. Global pooling methods use summation or neural networks to pool all the representations of nodes in each layer. Graphs with different structures can be processed because global pooling methods collect all the representations.
\citeauthor{DBLP:journals/corr/GilmerSRVD17} viewed GNNs as message passing schemes, and proposed a general framework for graph classification where entire graph representations could be obtained by utilizing the Set2Set\cite{SET2SET} method. SortPool\cite{zhang2018end} sorts embeddings for nodes according to the structural roles of a graph and feeds the sorted embeddings to the next layers.

\textbf{Hierarchical pooling}
Global pooling methods do not learn hierarchical representations which are crucial for capturing structural information of graphs. The main motivation of hierarchical pooling methods is to build a model that can learn feature- or topology-based node assignments in each layer. 
\citeauthor{RexYing} proposed DiffPool which is a differentiable graph pooling method that can learn assignment matrices in an end-to-end fashion. A learned assignment matrix in layer $l$, $S^{(l)} \in \mathbb{R}^{n_l \times n_{l+1}}$ contains the probability values of nodes in layer $l$ being assigned to clusters in the next layer $l+1$. Here, $n_l$ denotes the number of nodes in layer $l$. Specifically, nodes are assigned by the following equation:
\begin{equation*}
    S^{(l)} = \text{softmax}(\text{GNN}_{l}(A^{(l)},X^{(l)}))
\end{equation*}
\begin{equation}
\label{eq:diffpool}
    A^{(l+1)} = S^{(l)\top}A^{(l)}S^{(l)}
\end{equation}
where $X$ denotes the node feature matrix and $A$ is the adjacency matrix.

\begin{figure*}[ht]
\vskip 0.2in
\begin{center}
\centerline{\includegraphics[width=0.95\textwidth]{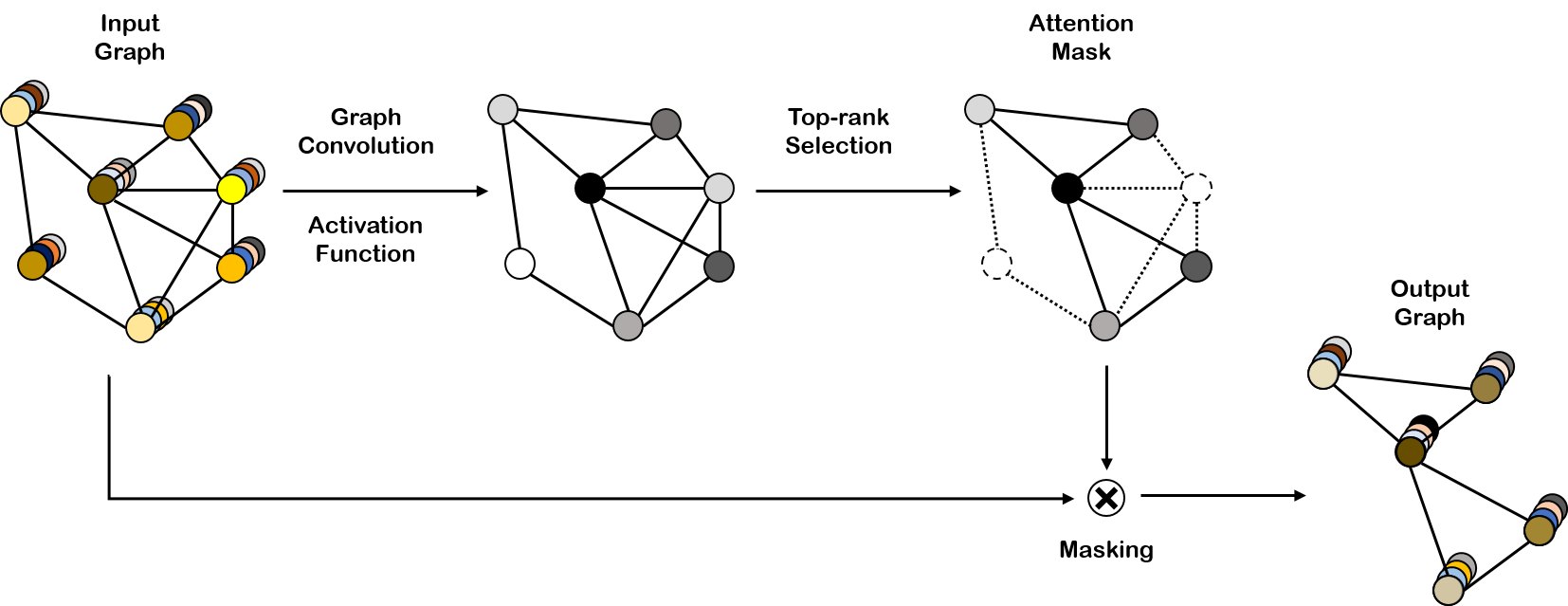}}
\caption{An illustration of the SAGPool layer.}
\label{SAGPool}
\end{center}
\vskip -0.2in
\end{figure*}

\citeauthor{cangea2018towards} utilized gPool\cite{gao2019graph} and achieved performance comparable to that of DiffPool. gPool requires a storage complexity of $\mathcal{O}(|V|+|E|)$ whereas DiffPool requires $\mathcal{O}(k|V|^2)$  where $V$, $E$, and $k$ denote vertices, edges, and pooling ratio, respectively. gPool uses a learnable vector $p$ to calculate projection scores, and then uses the scores to select the top ranked nodes. Projection scores are obtained by the dot product between $p$ and the features of all the nodes. The scores indicate the amount of information of nodes that can be retained. 
The following equation roughly describes the pooling procedure in gPool.
\begin{equation*}
    y = X^{(l)}\mathbf{p}^{(l)}/\|\mathbf{p}^{(l)}\|, \quad \text{idx} = \text{top-rank}(y,\lceil kN\rceil)
\end{equation*}
\begin{equation}
    \label{eq:gpool}
    A^{(l+1)} = A^{(l)}_{\text{idx},\text{idx}}
\end{equation}
As in Equation (\ref{eq:gpool}), the graph topology does not affect the projection scores.

To further improve graph pooling, we propose SAGPool which can use features and topology to yield hierarchical representations with a reasonable complexity of time and space.

%% file: sections/3_proposed_method.tex
\section{Proposed Method}
\label{method}
The key point of SAGPool is that it uses a GNN to provide self-attention scores. In Section \ref{method:SAGP}, we describe the mechanism of SAGPool and its variants. Model architectures for the evaluations are described in Section \ref{method:architecture}. The SAGPool layer and the model architectures are illustrated in Figure \ref{SAGPool} and Figure \ref{model}, respectively.

\subsection{Self-Attention Graph Pooling}
\label{method:SAGP}

\textbf{Self-attention mask} 
Attention mechanisms have been widely used in the recent deep learning studies \cite{parikh2016decomposable, cheng2016long, zhang2018self, veličković2018graph}. Such mechanisms make it possible to focus more on important features and less on unimportant features. In particular, self-attention, commonly referred to as intra-attention, allows input features to be the criteria for the attention itself \cite{vaswani2017attention}. We obtain self-attention scores using graph convolution. For instance, if the graph convolution formula of \citeauthor{kipf2016semi} is used, the self-attention score $Z \in \mathbb{R}^{N \times 1}$ is calculated as follows.
\begin{equation}
\label{eq:GCN_att}
    Z = \sigma(\tilde{D}^{-\frac{1}{2}}\tilde{A}\tilde{D}^{-\frac{1}{2}}X\Theta_{att})
\end{equation}
where $\sigma$ is the activation function (e.g. $tanh$), $\tilde{A} \in \mathbb{R}^{N \times N}$ is the adjacency matrix with self-connections (i.e. $\tilde{A} = A + I_N$), $\tilde{D} \in \mathbb{R}^{N \times N}$ is the degree matrix of $\tilde{A}$, $X \in \mathbb{R}^{N \times F}$ is the input features of the graph with $N$ nodes and $F$-dimensional features, and $\Theta_{att} \in \mathbb{R}^{F \times 1}$ is the only parameter of the SAGPool layer. By utilizing graph convolution to obtain self-attention scores, the result of the pooling is based on both graph features and topology. We adopt the node selection method of \citeauthor{gao2019graph, cangea2018towards}, which retains a portion of nodes of the input graph even when graphs of varying sizes and structures are inputted. The pooling ratio $k\in (0,1]$ is a hyperparameter that determines the number of nodes to keep. The top $\lceil kN\rceil$ nodes are selected based on the value of $Z$.
\begin{equation}
    \text{idx} = \text{top-rank} (Z,\lceil kN\rceil), \quad Z_{mask} = Z_{\text{idx}} 
\end{equation}
where $\text{top-rank}$ is the function that returns the indices of the top $\lceil kN \rceil $ values, $\cdot_{\text{idx}}$ is an indexing operation and  $Z_{mask}$ is the feature attention mask. 

\textbf{Graph pooling}
An input graph is processed by the operation notated as \textbf{masking} in Figure \ref{SAGPool}.
\begin{equation}
\label{eq:masking}
    X' = X_{\text{idx},:}, \; X_{out} = X' \odot Z_{mask}, \; A_{out} = A_{\text{idx},\text{idx}}
\end{equation}
where $X_{\text{idx},:}$ is the row-wise (i.e. node-wise) indexed feature matrix, $\odot$ is the broadcasted elementwise product, and $A_{\text{idx},\text{idx}}$ is the row-wise and col-wise indexed adjacency matrix. $X_{out}$ and $A_{out}$ are the new feature matrix and the corresponding adjacency matrix, respectively. 

\textbf{Variation of SAGPool}
The main reason for using graph convolution in SAGPool is to reflect the topology as well as node features. The various formulas of GNNs can be substituted for Equation (\ref{eq:GCN_att}), if GNNs take the node feature and the adjacency matrix as inputs. The generalized equation for calculating the attention score $Z \in \mathbb{R}^{N \times 1}$ is as follows.
\begin{equation}
    \label{eq:general_mask}
    Z = \sigma(\text{GNN}(X,A))
\end{equation}
where $X$ denotes the node feature matrix and $A$ is the adjacency matrix.

There are several ways to calculate attention scores using not only adjacent nodes but also multi-hop connected nodes. In Equation (\ref{eq:augmentation}) and (\ref{eq:serial_mask}), we illustrate examples of using the two-hop connections which involve the augmentation of edges and the stack of GNN layers. Adding the square of an adjacency matrix creates edges between two-hop neighbors. 
\begin{equation}
    \label{eq:augmentation}
    Z = \sigma(\text{GNN}(X,A + A^2))
\end{equation}
The stack of GNN layers allows for the indirect aggregation of two-hop nodes. In this case, the nonlinearity and the number of parameters of the SAGPool layer increase. 
\begin{equation}
    \label{eq:serial_mask}
    Z = \sigma(\text{GNN}_2(\sigma(\text{GNN}_1(X,A)),A))
\end{equation}
Equations (\ref{eq:augmentation}) and (\ref{eq:serial_mask}) can be applied to the multi-hop connections.

Another variant is to average multiple attention scores. The average attention score is obtained by $M$ GNNs as follows: 
\begin{equation}
    \label{eq:parallel_mask}
    Z = \frac{1}{M} \underset{m}{\sum} \sigma(\text{GNN}_m(X,A))
\end{equation}

In this paper, the models using Equation (\ref{eq:augmentation}), (\ref{eq:serial_mask}), and (\ref{eq:parallel_mask}) are referred to as SAGPool$_{\text{augmentation}}$, SAGPool$_{\text{serial}}$ , and SAGPool$_{\text{parallel}}$, respectively.

\subsection{Model Architecture}
\label{method:architecture}
According to \citeauthor{lipton2018troubling}, if numerous modifications are made to a model, it may be difficult to identify which modification contributes to improving performance. For a fair comparison, we adopted the model architectures from \citeauthor{zhang2018end} and \citeauthor{cangea2018towards}, and compared the baselines and our method using the same architectures.

\textbf{Convolution layer}
As mentioned in Section \ref{related:graphconvolution}, there are many definitions for graph convolution. Other types of graph convolution may improve performance, but we utilize the widely used graph convolution proposed by \citeauthor{kipf2016semi} for all the models. Equation (\ref{eq:GCN}) is the same as Equation (\ref{eq:GCN_att}), except for the dimension of $\Theta$.
\begin{equation}
\label{eq:GCN}
    h^{(l+1)} = \sigma(\tilde{D}^{-\frac{1}{2}}\tilde{A}\tilde{D}^{-\frac{1}{2}}h^{(l)}\Theta)
\end{equation}
where $h^{(l)}$ is the node representation of $l$-th layer and $\Theta \in \mathbb{R}^{F \times F'}$ is the convolution weight with input feature dimension $F$ and output feature dimension $F'$. The Rectified Linear Unit (ReLU) \cite{nair2010rectified} function is used as an activation function.

\textbf{Readout layer}
Inspired by the JK-net architecture \cite{xu2018representation}, \citeauthor{cangea2018towards} proposed a readout layer that aggregates node features to make a fixed size representation. The summarized output feature of the readout layer is as follows:
\begin{equation}
\label{eq:readout}
    s = \frac{1}{N} \underset{i=1}{\overset{N}{\sum}}x_i \  ||\  \underset{i=1}{\overset{N}{\max}}\ x_i
\end{equation}
where $N$ is the number of nodes, $x_i$ is the feature vector of $i$-th node, and $||$ denotes concatenation.

\begin{figure}[t!]
\vskip 0.2in
\begin{center}
\centerline{\includegraphics[width=\columnwidth]{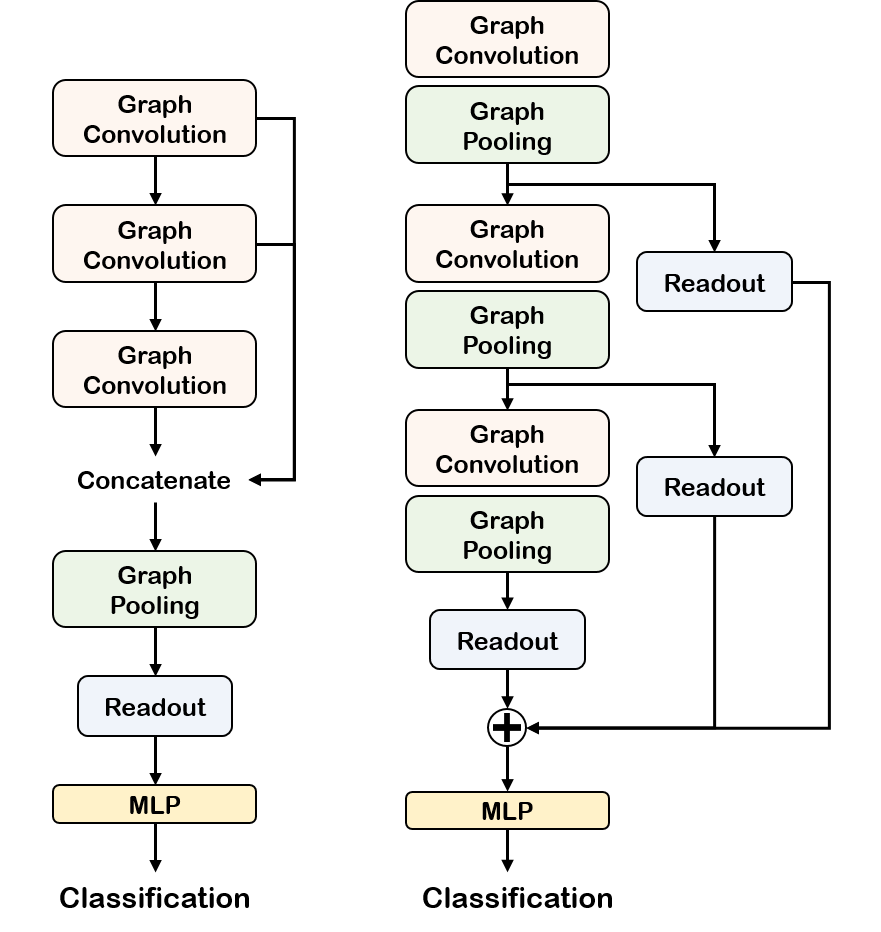}}
\caption{The global pooling architecture (left) and the hierarchical pooling architecture (right). These architectures are applied to all the baselines and SAGPool for a fair comparison. In this paper, the architecture on the left side is referred to as $POOL_g$ and the architecture on the right side is referred to as $POOL_h$ with the $POOL$ method (e.g. SAGPool$_g$, gPool$_h$).}
\label{model}
\end{center}
\vskip -0.2in
\end{figure}

\textbf{Global pooling architecture}
We implemented the global pooling architecture proposed by \citeauthor{zhang2018end}. As shown in Figure \ref{model}, the global pooling architecture consists of three graph convolutional layers and the outputs of each layer are concatenated. Node features are aggregated in the readout layer which follows the pooling layer. Then graph feature representations are passed to the linear layer for classification.

\begin{table*}[t!]
\caption{Statistics of data sets.}
\label{tab:statistics}
\vskip 0.15in
\begin{center}
\begin{small}
\begin{tabular}{lcccc}
\toprule
\textbf{Data set} & \textbf{Number of Graphs} & \textbf{Number of Classes} & \textbf{Avg. \# of Nodes per Graph} & \textbf{Avg. \# of Edges per Graph}\\
\midrule
D\&D    & 1178 & 2 & 284.32 & 715.66 \\
PROTEINS    & 1113 & 2 & 39.06 & 72.82 \\
NCI1    & 4110 & 2 & 29.87 & 32.30 \\
NCI109    & 4127 & 2 & 29.68 & 32.13 \\
FRANKENSTEIN    & 4337 & 2 & 16.90 & 17.88 \\
\bottomrule
\end{tabular}
\end{small}
\end{center}
\vskip -0.1in
\end{table*}

\textbf{Hierarchical pooling architecture}
In this setting, we implemented the hierarchical pooling architecture from the recent hierarchical pooling study of \citeauthor{cangea2018towards}. As shown in Figure \ref{model}, the architecture is comprised of three blocks each of which consists of a graph convolutional layer and a graph pooling layer. The outputs of each block are summarized in the readout layer. The summation of the outputs of each readout layer is fed to the linear layer for classification.

%% file: sections/4_experiments.tex
\section{Experiments}
\label{experiments}
We evaluate the global pooling and hierarchical pooling methods on the graph classification task. In Section \ref{ex:data}, we discuss the datasets used for evaluation. Section \ref{ex:train} describes how we train the models. The methods compared in the experiments are introduced in Sections \ref{ex:baseline} and \ref{ex:variation}.

\begin{table}
\caption{The grid search space for the hyperparameters. The pooling ratio is used only for the hierarchical pooling architecture because the the global pooling architecture uses the same node selection strategy as SortPool. The node selection strategy of SortPool does not require the pooling ratio.}
\label{tab:hyperparameter}
\vskip 0.15in
\begin{center}
\begin{small}
\begin{tabular}{lcc}
\toprule
\textbf{Hyperparameter} & \textbf{Range} \\

\midrule
Learning rate   & 1e-2, 5e-2, 1e-3, 5e-3, 1e-4, 5e-4 \\
\midrule
Hidden size  & 16, 32, 64, 128 \\
\midrule
Weight decay  & \multirow{2}{*}{1e-2, 1e-3, 1e-4, 1e-5}\\
(L2 regularization) & \\
\midrule
Pooling ratio & 1/2, 1/4 \\
\bottomrule
\end{tabular}
\end{small}
\end{center}
\vskip -0.1in
\end{table}

\subsection{Datasets}
\label{ex:data}
Five datasets with a large number of graphs ($ > 1$k) were selected among the benchmark datasets \cite{KKMMN2016}. The statistics of the datasets are summarized in Table \ref{tab:statistics}.

\textbf{D\&D} \cite{dobson2003distinguishing, shervashidze2011weisfeiler} contains graphs of protein structures. A node represents an amino acid and edges are constructed if the distance of two nodes is less than 6 $\textup{\AA}$. A label denotes whether a protein is an enzyme or non-enzyme. \textbf{PROTEINS} \cite{dobson2003distinguishing, borgwardt2005protein} is also a set of proteins, where nodes are secondary structure elements. If nodes have edges, the nodes are in an amino acid sequence or in a close 3D space. \textbf{NCI} \cite{wale2008comparison} is a biological dataset used for anticancer activity classification. In the dataset, each graph represents a  chemical compound, with nodes and edges representing atoms and chemical bonds, respectively. \textbf{NCI1} and \textbf{NCI109} are commonly used as benchmark datasets for graph classification. \textbf{FRANKENSTEIN} \cite{orsini2015graph} is a set of molecular graphs \cite{costa2010fast} with node features containing continuous values. A label denotes whether a molecule is a mutagen or non-mutagen.

\begin{table*}
\caption{Average accuracy and standard deviation of the 20 random seeds. The subscript $g$ (e.g. $POOL_g$) denotes the global pooling architecture and the subscript $h$ (e.g. $POOL_h$) denotes the hierarchical pooling architecture.}
\label{tab:results}
\vskip 0.15in
\begin{center}
\begin{small}
\begin{tabular}{lccccc}
\toprule
\textbf{Models} & \textbf{D\&D} & \textbf{PROTEINS} & \textbf{NCI1} & \textbf{NCI109} & \textbf{FRANKENSTEIN} \\

\midrule
Set2Set$_g$     & $71.27\pm 0.84$ & $66.06 \pm 1.66 $ & $68.55 \pm 1.92 $ & $ 69.78 \pm 1.16 $ & $61.92 \pm 0.73 $ \\
SortPool$_g$    & $72.53 \pm 1.19$ & $ 66.72 \pm 3.56 $ & $73.82 \pm 0.96 $ & $74.02 \pm 1.18 $ & $60.61 \pm 0.77 $ \\
SAGPool$_g$ (Ours)  & $\boldsymbol{76.19} \pm 0.94 $ & $ \boldsymbol{70.04} \pm 1.47 $ & $\boldsymbol{74.18} \pm 1.20 $ & $\boldsymbol{74.06} \pm 0.78 $ & $\boldsymbol{62.57} \pm 0.60 $  \\
\midrule
DiffPool$_h$            & $ 66.95 \pm 2.41 $ & $68.20 \pm 2.02 $ & $62.32 \pm 1.90 $ & $61.98 \pm 1.98 $ & $60.60 \pm 1.62 $  \\
gPool$_h$               & $75.01 \pm 0.86$ & $71.10 \pm 0.90$ & $67.02\pm 2.25 $ & $ 66.12 \pm 1.60 $ & $ 61.46 \pm 0.84 $  \\
SAGPool$_h$  (Ours)     & $\boldsymbol{76.45} \pm 0.97$ & $\boldsymbol{71.86} \pm 0.97$ & $\boldsymbol{67.45} \pm 1.11 $ & $ \boldsymbol{67.86} \pm 1.41 $ & $ \boldsymbol{61.73} \pm 0.76 $  \\
\bottomrule
\end{tabular}
\end{small}
\end{center}
\vskip -0.1in
\end{table*}

\begin{table}
\caption{Experimental results of SAGPool$_h$ variants. We compare  ChebConv(K=2) \cite{defferrard2016convolutional}, GCNConv \cite{kipf2016semi}, SAGEConv \cite{hamilton2017inductive}, and GATConv(heads=6) \cite{veličković2018graph}. GCNConv is applied to SAGPool$_h$, SAGPool$_h,_{\text{augmentation}}$, SAGPool$_h,_{\text{serial}}$, and SAGPool$_h,_{\text{parallel}}$.}
\label{tab:variance}
\vskip 0.15in
\begin{center}
\begin{small}
\begin{tabular}{lcc}
\toprule
\textbf{Graph Convolution} & \textbf{D\&D} & \textbf{PROTEINS} \\
\midrule
SAGPool$_h$               & $76.45 \pm 0.97$ & $71.86 \pm 0.97$  \\
\midrule
SAGPool$_h,_{\text{Cheb}}$              & $75.82 \pm 0.79 $ & $ 71.98 \pm 0.93 $  \\
SAGPool$_h,_{\text{SAGE}}$              & $76.28 \pm 1.06 $ & $71.93 \pm 0.82 $  \\
SAGPool$_h,_{\text{GAT}}$               & $75.49 \pm 0.93 $ & $ 71.98 \pm 1.01 $ \\
\midrule
SAGPool$_h,_{\text{augmentation}}$           & $77.07 \pm 0.82 $ & $71.82 \pm 0.81$  \\
SAGPool$_h,_{\text{serial,2layers}}$    & $ 76.68 \pm 0.96 $ & $72.17 \pm 0.87 $  \\
\midrule
SAGPool$_h,_{\text{parallel},M=2}$      & $75.79 \pm 0.96 $ & $ 72.05 \pm 0.43$  \\
SAGPool$_h,_{\text{parallel},M=4}$      & $76.77 \pm 0.61 $ & $71.66 \pm 0.98$ \\
\bottomrule
\end{tabular}
\end{small}
\end{center}
\vskip -0.1in
\end{table}

\subsection{Evaluation of GNNs}  In addition, the same early stopping criterion and hyperparameter selection strategy are used for all the models to ensure a fair comparison.

\subsection{Training Procedures}
\label{ex:train}
\citeauthor{DBLP:journals/corr/abs-1811-05868} demonstrate that different splits of data can affect the performance of GNN models. In our experiments, we evaluated the pooling methods over 20 random seeds using 10-fold cross validation. A total of 200 testing results were used to obtain the final accuracy of each method on each dataset. 10 percent of the training data was used for validation in the training session. We used the Adam optimizer \cite{DBLP:journals/corr/KingmaB14}, early stopping criterion, patience, and hyperparameter selection strategy for the global pooling architecture and hierarchical pooling architecture. We stopped the training if the validation loss did not improve for 50 epochs in an epoch termination condition with a maximum of 100k epochs, as done in \cite{DBLP:journals/corr/abs-1811-05868}. The optimal hyperparameters are obtained by grid search. The ranges of grid search are summarized in Table \ref{tab:hyperparameter}.

\subsection{Baselines}
\label{ex:baseline}
We consider the following four pooling methods as baselines: Set2Set, SortPool, DiffPool, and gPool. DiffPool, gPool, and SAGPool$_h$ were compared using the hierarchical pooling architecture while Set2Set, SortPool, and SAGPool$_g$ were compared using the global pooling architecture. We used the same hyperparameter search strategy for all the baselines and SAGPool. The hyperparameters are summarized in Table \ref{tab:hyperparameter}.

\textbf{Set2Set} \cite{SET2SET} requires an additional hyperparameter which is the number of processing steps for the LSTM\cite{lstm} module. We use 10 processing steps for all the experiments. We assume that the readout layer is unnecessary because the LSTM module produces embeddings for graphs invariant to the order of nodes.

\textbf{SortPool} \cite{zhang2018end} is a recent global pooling method which uses sorting for pooling. The $K$ number of nodes is set such that 60\% of graphs have more than $K$ nodes. In the global pooling setting, SAGPool$_g$ has the same $K$ number of output nodes as SortPool.

\textbf{DiffPool} \cite{RexYing} is the first end-to-end trainable graph pooling method that can produce hierarchical representations of graphs. We did not use batch normalization for DiffPool, which is not related to the pooling method. For the hyperparameter search, the pooling ratio ranges from 0.25 to 0.5 for the following reasons. In the reference implementation, the cluster size is set to 25\% of the maximum number of nodes. DiffPool$_h$ causes the out of memory error when the pooling ratio is larger  than 0.5.

\textbf{gPool} \cite{gao2019graph} selects top-ranked nodes for pooling, which makes it similar to our method. The comparison between our method and gPool demonstrates that considering topology can help improve performance on the graph classification task.

\subsection{Variations of SAGPool}
\label{ex:variation}
As mentioned in section \ref{method:SAGP}, three variations of SAGPool are used to obtain attention scores $Z$. In our experiments, we compared each variant on the two datasets. First, any kind of GNNs can be applied to Equation (\ref{eq:general_mask}). We compared the performance of the three most widely used GNNs (SAGPool$_{\text{Cheb}}$, SAGPool$_{\text{SAGE}}$, SAGPool$_{\text{GAT}}$).
Second, we made the following modifications to SAGPool so that it can consider the two-hop connection: an edge augmentation (SAGPool$_{\text{augmentation}}$) in Equation (\ref{eq:augmentation}) and a stack of GNN layers (SAGPool$_{\text{serial}}$) in Equation (\ref{eq:serial_mask}). Last, multiple GNNs calculate attention scores and the scores are averaged to obtain the final attention score (SAGPool$_{\text{parallel}}$). We evaluated the performance of $M=2$ and $M=4$ using Equation (\ref{eq:parallel_mask}). The results are summarized in Table \ref{tab:variance}.

\begin{figure}[t!]
\vskip 0.2in
\begin{center}
\centerline{\includegraphics[width=\columnwidth]{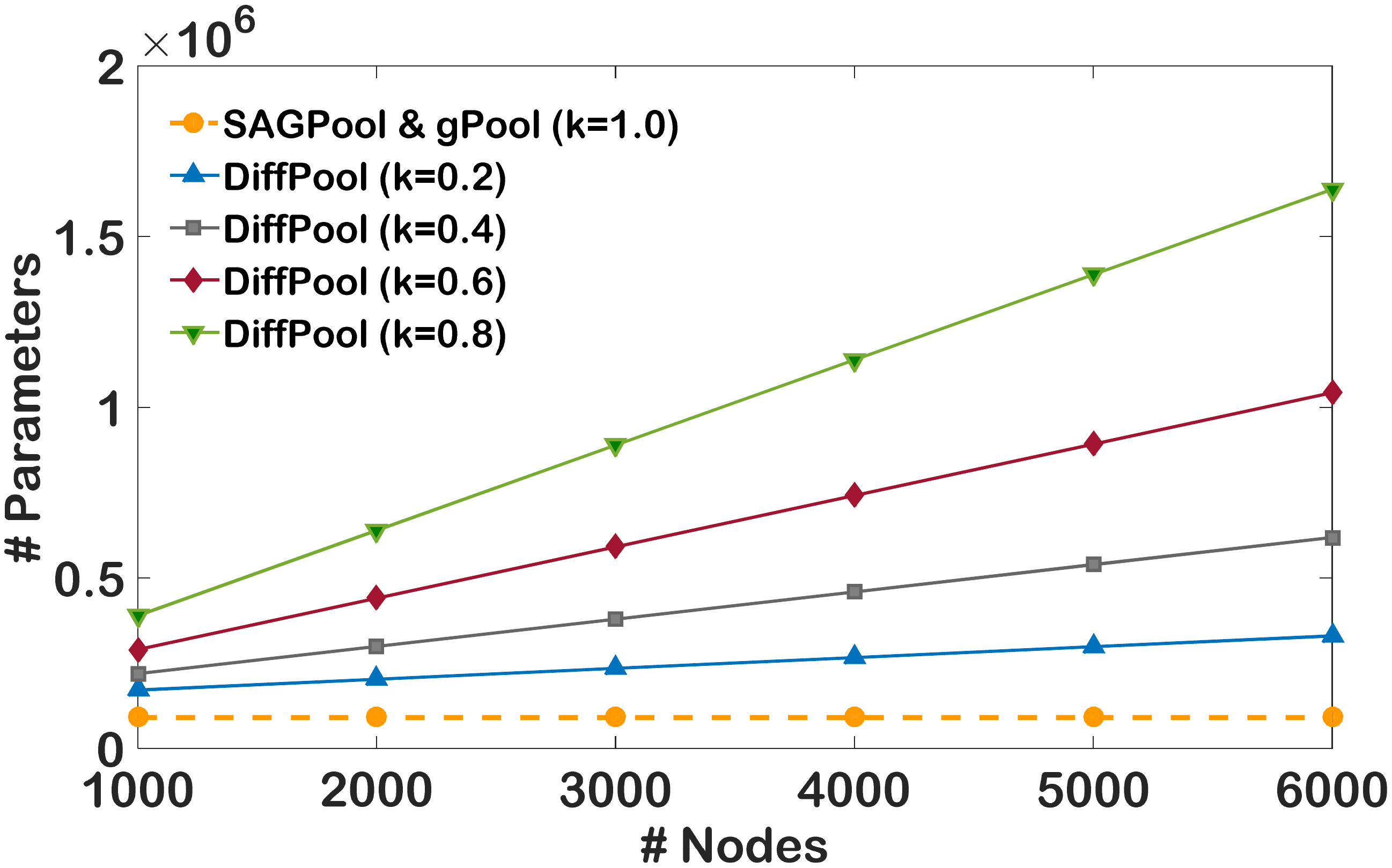}}
\caption{The increase in the number of parameters according to the number of graph nodes. The $x$-axis label denotes the number of input graph nodes and the $y$-axis label denotes the number of parameters of the hierarchical pooling models: the number of input node features is 128, the hidden feature size is 128, and the number of classes is 2. Equation (\ref{eq:GCN_att}) is used as a graph convolution of SAGPool. $k$ denotes the pooling ratio and $k=1.0$ indicates that the entire node is preserved after pooling. gPool and SAGPool have a consistent number of parameters regardless of the input graph size and the pooling ratio.}
\label{plot}
\end{center}
\vskip -0.2in
\end{figure}

\subsection{Summary of Results}
\label{ex:summary}
The results are summarized in Table \ref{tab:results} and \ref{tab:variance}. The accuracies and standard deviations are given in percentages. From the comparison of the global pooling methods and SAGPool, the results demonstrate that SAGPool generally performs well, but it performs especially well on D\&D and PROTEINS. In the experiments, SAGPool outperformed the hierarchical pooling methods on all the datasets. We also compared variants of SAGPool with the hierarchical pooling architecture on the two benchmark datasets. The performance of the variants of SAGPool varied. The experimental results of the SAGPool variants show that SAGPool has the potential to improve performance. A detailed analysis of the experimental results is provided in the next section.

%% file: sections/5_analysis.tex
\section{Analysis}
\label{analysis}
In this section, we provide an analysis of the experimental results. In Section \ref{analysis:global-hierarhcical}, we compare global pooling and hierarchical pooling. Section \ref{analysis:effect-topology} provides an explanation on how the SAGPool method addresses the shortcomings of the gPool method. In the \ref{analysis:sparsity} and \ref{analysis:relation-num-nodes} sections, we compare the efficiency of SAGPool with that of DiffPool. We provide an analysis of SAGPool variants in Section \ref{analysis:variants}.

\subsection{Global and Hierarchical Pooling}
\label{analysis:global-hierarhcical}
It is difficult to determine whether the global pooling architecture or hierarchical pooling architecture is completely beneficial to graph classification. Since the global pooling architecture $POOL_g$ (SAGPool$_g$, SortPool$_g$, Set2Set$_g$) minimizes the loss of information, it  performs better than the hierarchical pooling architecture $POOL_h$ (SAGPool$_h$, gPool$_h$, DiffPool$_h$) on datasets with fewer nodes (NCI1, NCI109, FRANKENSTEIN).  However, $POOL_h$ is more effective on datasets with a large number of nodes (D\&D, PROTEINS) because it efficiently extracts useful information from large scale graphs. Therefore, it is important to use the  pooling architecture that is the most suitable for the given data. Nonetheless, SAGPool tends to perform well with each architecture.

\subsection{Effect of Considering Graph Topology}
\label{analysis:effect-topology}
To calculate the attention scores of nodes, SAGPool$_h$ utilizes the graph convolution in Equation (\ref{eq:GCN_att}). Unlike gPool, SAGPool uses the $\tilde{D}^{-\frac{1}{2}}\tilde{A}\tilde{D}^{-\frac{1}{2}}$ term, which is the first order approximation of the graph Laplacian. This term allows SAGPool to consider graph topology. As shown in Table \ref{tab:results}, considering graph topology improves performance. In addition, the graph Laplacian does not have to be recalculated because it is the term used in a previous graph convolutional layer in the same block. Although SAGPool has the same parameters as gPool (Figure \ref{plot}), it achieves superior performance in the graph classification task.

\subsection{Sparse Implementation}
\label{analysis:sparsity}
Manipulating graph data with a sparse matrix is important for GNNs because the adjacency matrix is usually sparse. When graph convolution is calculated using a dense matrix, the computational complexity of multiplication $AX$ is $\mathcal{O}(|V|^2)$ where $A$ is the adjacency matrix, $X$ is the feature matrix of nodes, and $V$ denotes vertices. Pooling with a dense matrix causes the memory efficiency problem, as mentioned by \cite{cangea2018towards}. However, if a sparse matrix is used in the same operation, the computational complexity is reduced to $\mathcal{O}(|E|)$ where $E$ represents the edges. Since SAGPool is a sparse pooling method, it can reduce its computational complexity, unlike DiffPool which is a dense pooling method. 
Sparseness also affects space complexity. Since SAGPool uses GNN for obtaining attention scores, SAGPool requires 
$\mathcal{O}(|V|+|E|)$ of storage for sparse pooling whereas dense pooling methods need $\mathcal{O}(|V|^2)$.

\subsection{Relation with the Number of Nodes}
\label{analysis:relation-num-nodes}
In DiffPool, the cluster size has to be defined when constructing a model because a GNN produces an assignment matrix $S$ as stated in Equation (\ref{eq:diffpool}). The cluster size has to be proportional to the maximum number of nodes according to the reference implementation. These requirements of DiffPool can lead to two problems. 
First, the number of parameters is dependent on the maximum number of nodes as shown in Figure \ref{plot}. Second, it is difficult to determine the right cluster size when the number of nodes varies greatly. For example, only 10 out of 1178 graphs have over 1000 nodes, where the maximum number of nodes is 5748 and the minimum is 30. The cluster size is 574 if the pooling ratio is 10\%, which expands the size of graphs after pooling for most of the data. On the other hand, in SAGPool, the number of parameters is independent of the cluster size. In addition, the cluster size can be changed based on the number of input nodes.


\subsection{Comparison of the SAGPool Variants}
\label{analysis:variants}
To investigate the potential of our method, we evaluated SAGPool variants on two datasets. SAGPool can be modified to perform the following: changing the type of GNN, considering the two-hop connections, and averaging the attention scores of multiple GNNs. As shown in Table \ref{tab:variance}, the performance on graph classification varies depending on which dataset and type of GNN in SAGPool are used.
We used two techniques to consider two-hop connections. The attention scores obtained by the two sequential GNN layers (SAGPool$_{\text{serial}}$) reflect the information of two-hop neighbors. 
Another technique is to add the square of an adjacency matrix to itself, resulting in a new adjacency matrix that has two-hop connectivity. Without any modifications to the SAGPool layer, the new adjacency matrix can be processed in SAGPool$_{\text{augmentation}}$. The information of two-hop neighbors may help improve performance.
The last variants of SAGPool is to average the attention scores from multiple GNNs. We found that choosing the right $M$ for the dataset can help achieve stable performance.

\subsection{Limitations}
\label{analysis:limit}
We retain a certain percentage (pooling ratio $k$) of nodes to handle different input graphs of various sizes, which has also been done in previous studies \cite{gao2019graph,cangea2018towards}. 
In SAGPool, we cannot parameterize the pooling ratios to find optimal values for each graph. To address this limitation, we used binary classification to decide which nodes to preserve, but this did not completely solve the issue.

%% file: sections/6_conclusion.tex
\section{Conclusion}
\label{conclusion}
In this paper, we proposed SAGPool which is a novel graph pooling method based on self-attention. Our method has the following features: hierarchical pooling, consideration of both node features and graph topology, reasonable complexity, and end-to-end representation learning. SAGPool uses a consistent number of parameters regardless of the input graph size. 
Extensions of our work may include using learnable pooling ratios to obtain optimal cluster sizes for each graph and studying the effects of multiple attention masks in each pooling layer, where final representations can be derived by aggregating different hierarchical representations.
Our experiments were run on a NVIDIA TitanXp GPU. We implemented all the baselines and SAGPool using PyTorch \cite{paszke2017automatic} and the geometric deep learning extension library provided by \citeauthor{Fey/etal/2018}.